\documentclass[journal]{IEEEtran}
\usepackage{amsfonts}
\usepackage{amssymb}
\usepackage{amsthm}
\usepackage{float}
\usepackage{graphicx}
\usepackage{amssymb,amsthm} 
\usepackage{mathtools}
\usepackage{amsmath, amsfonts,epsfig, multirow, floatflt}
\usepackage{epsfig,latexsym,amsfonts,amssymb,amsmath,verbatim}
\usepackage{subfigure}
\usepackage{mathrsfs}
\usepackage{cite}
\usepackage{color}
\usepackage{soul}
\usepackage{multirow}
\usepackage{multicol}
\usepackage{array}
\usepackage{algorithm}
\usepackage{algpseudocode}
\usepackage{setspace}
\usepackage{booktabs}
\usepackage{threeparttable}
\usepackage{makecell}
\usepackage{epstopdf}
\usepackage{tabularx}
\usepackage{stfloats}
\usepackage{geometry}

\newcounter{MYalgorithmic}
\renewcommand{\theMYalgorithmic}{\arabic{MYalgorithmic}}
\newcommand{\algcaption}[1]{
\refstepcounter{MYalgorithmic}
\textbf{Algorithm}~\textbf{\theMYalgorithmic}.~#1}
\newenvironment{MYalgorithmic}[5]
{
\hrule height 1.2pt
\vspace{3pt}
#1{#2}%
#3{#4}
\vspace{3pt}
\hrule height 0.5pt
\vspace{3pt}
#5
}
{
\vspace{3pt}
\hrule height 0.5pt
}
\graphicspath{{figure/}}
\usepackage{caption}
\ifCLASSINFOpdf
\else
\fi
\newtagform{brackets}{(}{)}
\usetagform{brackets}

\begin{document}
\title{A Driving Intention Prediction Method Based on Hidden Markov Model for Autonomous Driving}
\author{\IEEEauthorblockN{Shiwen Liu, Kan Zheng, \textit{Senior Member, IEEE}, Long Zhao, \textit{Member, IEEE}, and Pingzhi Fan, \textit{Fellow, IEEE}}
}
\maketitle

\begin{abstract}
In a mixed-traffic scenario where both autonomous vehicles and human-driving vehicles exist, a timely prediction of driving intentions of nearby human-driving vehicles is essential for the safe and efficient driving of an autonomous vehicle. In this paper, a driving intention prediction method based on Hidden Markov Model (HMM) is proposed for autonomous vehicles. HMMs representing different driving intentions are trained and tested with field collected data from a flyover. When training the models, either discrete or continuous characterization of the mobility features of vehicles is applied. Experimental results show that the HMMs trained with the continuous characterization of mobility features can give a higher prediction accuracy when they are used for predicting driving intentions. Moreover, when the surrounding traffic of the vehicle is taken into account, the performances of the proposed prediction method are further improved.
\end{abstract}
\begin{IEEEkeywords}
driving intention prediction, autonomous driving, hidden Markov model.
\end{IEEEkeywords}

\section{Introduction}
Traffic safety has always been one of the important issues in human society. With the development of autonomous driving technology, a mixed-traffic urban environment is arising, in which autonomous vehicles have to interact with human-driven vehicles \cite{5}. In the interaction between autonomous and human-driven vehicles, how to avoid traffic accidents caused by unmanned driving has become a research field of great concern \cite{16}. Generally, autonomous vehicles have to make decisions in dynamic and uncertain environments. The uncertainty comes from the fact that the intention of human drivers cannot be directly measured \cite{10}. Hence, for an unmanned vehicle, the accurate prediction of the expected behavior of other vehicles is essential to avoid the threat of traffic accidents. In the traditional traffic scenario where only human-driven vehicles exist, human drivers can judge the moving intentions of the surrounding vehicles according to the established traffic rules and their driving experience. Based on the judgment, each driver adjusts his/her driving in real time, in order to ensure the safety and efficiency of the traffic. However, in the mixed-traffic scenario, the unmanned vehicles have to estimate the driving intentions of the human-driven vehicles on the road based on pre-established prediction models. For an unmanned vehicle, it can obtain the driving status of another vehicle on the road based on communication techniques such as vehicle-to-vehicle (V2V) and vehicle-to-infrastructure (V2I) communications. Research on the communication techniques in vehicular networks has been widely carried out \cite{12}-\cite{14}, which can guarantee reliable communications among unmanned vehicles and human-driven vehicles. With the driving status of nearby vehicles obtained via communications, an unmanned vehicle can apply the pre-established prediction model to predict the future driving intentions of the nearby vehicles. \par
In recent years, researchers have been working on the recognition and  prediction of driving intentions of vehicles. For example, Bayesian decision, support vector machine (SVM), and hidden Markov model (HMM) etc., are widely used. In \cite{9}, the authors proposed an algorithm to predict driver's intention with fuzzy logic and edit distance. In \cite{2} and \cite{11}, SVM is implemented for driving intention recognition. The model for detecting cognitive distraction is developed using drivers’ eye movements and driving performance data. HMMs are applicable in characterizing the underlying relationship between observations and the hidden states that generate the observations. The authors in \cite{3} proposed a method of modeling driving behavior concerned with certain period of past movements by using AR-HMM, in order to predict  the stop probability of a vehicle. The methods developed in \cite{1} can be applied in ADAS to take appropriate measures in reducing accidents. The driver intention close to a road intersection is estimated, using discrete HMMs and the Hybrid State System (HSS) framework as basis. The driver decisions are depicted as a discrete state system at a higher level and the continuous vehicle dynamics are depicted as a continuous state system at a lower level in the HSS framework. The study in \cite{4} focuses on the scenario when vehicles merge because of reduction in the number of lanes on city roads, and considers the mutual interaction between drivers. However, the mobility data of vehicles used in most existing work is generated by driving simulators, which can not accurately reflect the driving conditions of the vehicle in real traffic environment. \par
In our work, the traffic data is collected from vehicles on a real road. When predicting the driving intention of a vehicle on this road, its historical movement trajectory is first considered, and then the surrounding traffic close to it is taken into account. Firstly, we use the collected data to train the prediction model based on HMM. Then, we use the trained model to predict the driving intention of a given vehicle. In this paper, the given vehicle is considered as the targeted vehicle, and the vehicles nearby are considered as the surrounding vehicles. When a trail of historical information of the targeted vehicle, or information of both the targeted vehicle and the surrounding vehicles is available, the most likely future driving intention of the targeted vehicle can be achieved through the proposed method. Moreover, either discrete or continuous characterization of this mobility information is applied to make the mobility features be used as observations in HMMs. In the discrete characterization, $K$-means clustering is used for discretizing the mobility data of vehicles. In the continuous characterization, the continuous mobility features are modeled as Gaussian mixture models (GMMs). \par
The main contributions of this paper are summarized as follows.
\begin{itemize}
	\item
	A driving intention prediction method is proposed based on HMM, which can be used to predict the future moving intention of a given targeted vehicle, when a trail of mobility features is available.
	\item
	The HMMs are trained with data collected from the vehicles on a real road, and can be better adapted to the real traffic environment.
	\item
	The prediction is carried out in the case where only the targeted vehicle is involved in, and in the case where both the targeted vehicle and the surrounding vehicles are involved in. When the mobility features of the surrounding vehicles are introduced, the performances of the proposed prediction method are further improved.

\end{itemize} \par
The rest of this paper is organized as follows. In Section \uppercase\expandafter{\romannumeral2}, we propose a driving intention prediction method based on HMM. The experiment scenario and numerical results are given in Section \uppercase\expandafter{\romannumeral3}. Section \uppercase\expandafter{\romannumeral4} concludes this paper. \par

\section{Prediction of driving intentions based on HMM}
\begin{figure}[!b]
	\centering
	\includegraphics[width=0.48\textwidth]{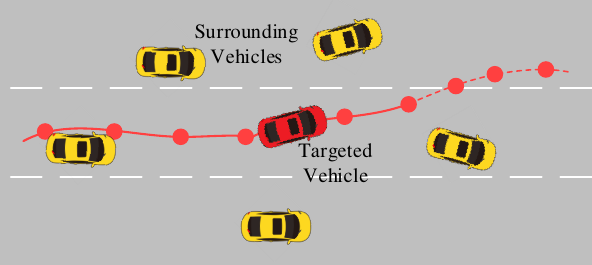}
	\caption{The targeted vehicle and the surrounding vehicles.}
\end{figure}
\begin{figure*}[!t]
	\centering
	\includegraphics[scale=0.55]{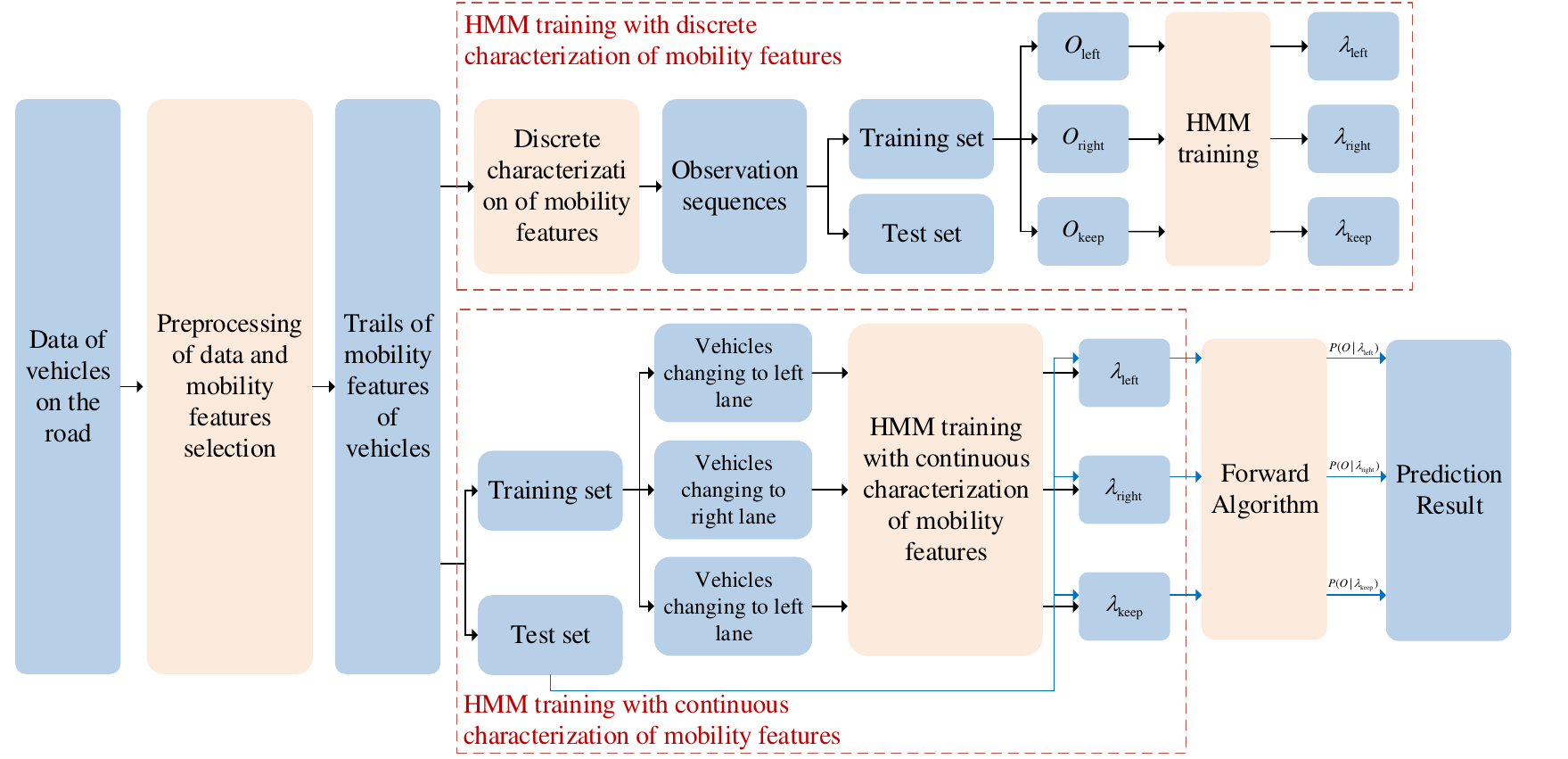}
	\caption{The framework of the proposed driving intention prediction method.}
\end{figure*}
In this section, we propose a driving intention prediction method based on HMM. The vehicle whose driving intention is required to be predicted is referred as the targeted vehicle, and the vehicles close to it are referred as surrounding vehicles, as shown in Fig.1. The process of the proposed method is illustrated in Fig.2. The trails of mobility features of vehicles are obtained, and the training of HMMs can be implemented in one of the two approaches. On one hand, in the HMM training with the discrete characterization of mobility features, all the trails of features are firstly turned into observation sequences. After that, they are divided into a training set and a test set. The vehicles in the training set are classified into three subsets, according to the driving intentions. Finally, HMMs representing different driving intentions are trained. On the other hand, in the HMM training with the continuous characterization of mobility features, the training set is firstly divided into subsets, and then the continuous characterization and the training of HMMs are processed at the same time. After the HMMs are well trained with one of the two approaches, vehicles in the test set can be used to test the models. In each experiment, the driving intention of one vehicle from the test set is predicted. The prediction accuracy is statistically measured among all experiments. The following subsections describe each step in detail.

\subsection{Mobility features of vehicles}
Information from the targeted vehicle and the surrounding vehicles can be used as features for HMM training and prediction. Firstly, the dataset provides dynamic locations of the vehicles on a selected road. For a particular vehicle, its location coordinates are recorded every certain seconds, and a successive trail of its locations is available. Moreover, in another dataset, the lanes of the roads are divided into segmented links, and the location coordinates of each link are provided. Then, by preprocessing the raw data, some types of mobility features such as velocity, acceleration of the vehicles, and the offsets between the vehicles and the lanes can be obtained. Finally, several types of mobility features are selected and used in HMM training and prediction. \par
In this paper, $N$ types of mobility features are selected for each vehicle in the HMM training and prediction. The trail of the $n$-th type of feature is denoted as a vector, i.e., 
\begin{equation}
{{\bf{x}}_n} = [{x_{n,1}},...,{x_{n,t}},...,{x_{n,T}}],
\end{equation}
where $t = 1,2,...,T$ is the index of the time step. It is assumed that the trails of all types of features are truncated to the same length of $T$ time steps. Then, the set of all types of features of this vehicle can be written as a matrix, i.e.,
\begin{equation}
{\bf{X}} = {\rm{[}}{\bf{x}}_1^{\mathsf{T}},...,{\bf{x}}_n^{\mathsf{T}},...,{\bf{x}}_N^{\mathsf{T}}],
\end{equation}
where $n{\rm{ = 1,2,}}...{\rm{,}}N$.\par
Note that ${\bf{X}}$ defined above is a matrix representing for a set of features for one vehicle. 
Generally, to make an HMM converged in the training, an adequate number of samples are required. Assume that $L$ vehicles are used to train an HMM, and let ${\bf{X}}^l$ denote the set of features of the $l$-th vehicle. Thus, a matrix ${\bf{F}} = {[{{\bf{X}}^1};...;{{\bf{X}}^{l}};...;{{\bf{X}}^{L}}]}$ has a dimension of $TL \times N$, and it is the matrix that includes the mobility features of all $L$ vehicles, namely mobility feature matrix. Note that the mobility feature matrix can include the mobility features of both the targeted vehicle and the surrounding vehicles. \par

\subsection{HMM training with discrete characterization of mobility features}
The mobility features can be represented in a discrete form by the technique of clustering. One possible approach is to apply $K$-means clustering, which is an exclusive clustering method based on distance. It classifies the sets of mobility features into $K$ clusters via unsupervised machine learning technique, and outputs the index of the cluster that each set belongs to \cite{6}\cite{17}. \par
As discussed above, the mobility features of $L$ vehicles are involved, and each vehicle gives a set of features at each time step. Thus, there are $TL$ sets of features, and each set $c_d$ is the $d$-th row of mobility feature matrix ${\bf{F}}$, where $d=1,2,...,TL$. Each $c_d$ is regarded as a data point in the $N$-dimensional space, and should be classified into one of the $K$ clusters in the space. The main idea is to minimize the sum of distance from the center points to the data points in the clusters. Let ${\mu _k}$ represent for the center point of clsuter $k$, the sum of distance is denoted as
\begin{equation}
J = \sum\limits_{d = 1}^{TL} {\sum\limits_{k = 1}^K {{r_{d,k}}\left\| {{c_d} - {\mu _k}} \right\|} },
\end{equation}
where $r_{d,k} = 1$ if $c_d$ is classified into the $k$-th cluster and $r_{d,k} = 0$ otherwise. 
To ensure that $J$ is minimized, ${\mu _k}$ should meet
\begin{equation}
{\mu _k} = \frac{{\sum\limits_{d = 1}^{TL} {{r_{d,k}}{c_d}} }}{{\sum\limits_{d = 1}^{TL} {{r_{d,k}}} }},
\end{equation}
where $k=1,2,...,K$. The detailed procedure of discrete characterization of mobility features by $K$-means clustering is described in Algorithm 1. \par 
\begin{figure}[!b]
	\begin{MYalgorithmic}
		\algcaption{Discrete characterization of mobility features by $K$-means clustering}
		\label{alg1}
		\begin{algorithmic}[1]
			\renewcommand{\algorithmicrequire}{\textbf{Input:}}
			\Require
			mobility feature matrix ${\bf{F}}$, number of clusters $K$.
			\renewcommand{\algorithmicrequire}{\textbf{Output:}}
			\Require
			observation sequences of $L$ vehicles $O = \{ {O^1},...,{O^l},...,{O^L}\}$.
			\renewcommand{\algorithmicrequire}{\textbf{Initialization:}}
			\Require
			\State Set initial values of the center points of the clusters: ${\mu _1, \mu _2, ..., \mu _K}$.
			\renewcommand{\algorithmicrequire}{\textbf{Step 1:}}
			\Require
			\State For any data point $c_d$, classify it into cluster $k$, if the center point of the cluskter $k$ is the nearest one of all $K$ center points to it.
			\renewcommand{\algorithmicrequire}{\textbf{Step 2:}}
			\Require
			\State Update ${r_{d,k}}$ according to the classification result;
			\State Update the values of center points ${\mu _k}$ by (4).
			\renewcommand{\algorithmicrequire}{\textbf{Step 3:}}
			\Require
			\State \textbf{IF} ${\mu _k}$ converges, denote the final classification result as $y(d)=k$ when ${r_{d,k}=1}$; The observation of each vehicle $O^l$ is obtained by $o_t^l = y[(l - 1)T + t]$, where $o_t^l$ is the element of $O^l$ at time step $t$; \\
			\textbf{ELSE} Return to \textbf{Step 1}.
		\end{algorithmic}
	\end{MYalgorithmic}
\end{figure}
In this paper, one particular HMM ${\lambda _i}$ is trained for each type of driving intention $i = 1,2,...,I$, where $I$ is the number of types of driving intention. For an HMM ${\lambda }$ (the index $i$ is omitted for simplicity), it includes a set of hidden states $H$, a set of observations $V$, state transition probabilities $A = \{ {a_{q,p}}\} $, state-observation probabilities $B = \{ {b_q}(j)\} $, and initial state probabilities $\pi  = \{ {\pi _q}\} $ \cite{7}\cite{15}. It can be represented as
\begin{equation}
\lambda  = \{ H,V,A,B,\pi \}.
\end{equation}
It is assumed that there are $Q$ possible hidden states in the set $H$. The hidden states might be the inside operations by the drivers that causes changes in the observations. \par
Given an HMM, the forward probability ${\alpha _t}(q)$ is defined as the probability of observing $o_1^l,o_2^l,...,o_t^l$ and the state of the Markov chain at time $t$ being the $q$-th state in the $H$, i.e.,
\begin{equation}
{\alpha _t}(q) = P(o_1^l,o_2^l,...,o_t^l,{s_t} = {H(q)}|\lambda ).
\end{equation}  
Similalry, the backward probability ${\beta _t}(q)$ is defined as
\begin{equation}
{\beta _t}(q) = P(o_{t+1}^l,o_{t+2}^l,...,o_T^l,{s_t} = {H(q)}|\lambda ).
\end{equation}
Then, the probability of the state at time $t$ being ${H(q)}$ is
\begin{equation}
{\eta _t}(q) = \frac{{{\alpha _t}(q){\beta _t}(p)}}{{\sum\limits_{p = 1}^Q {{\alpha _t}(p){\beta _t}(p)} }}.
\end{equation}
The probability of the state at time $t$ being ${H(q)}$ and the state at time $t+1$ being ${H(p)}$ can be obtained, i.e.,
\begin{figure}[!b]
	\begin{MYalgorithmic}
		\algcaption{HMM training with discrete characterization of mobility features}
		\label{alg2}
		\begin{algorithmic}[1]
			\renewcommand{\algorithmicrequire}{\textbf{Input:}}
			\Require
			observation sequences of $L$ vehicles $O=\{{O^1},...,{O^l},...,{O^L}\}$.
			\renewcommand{\algorithmicrequire}{\textbf{Output:}}
			\Require
			HMM parameters $A,B,\pi$.
			\renewcommand{\algorithmicrequire}{\textbf{Initialization:}}
			\Require
			\State Initial guess of $A,B,\pi$: $\tilde a_{q,p}^{(0)},\tilde b_p(j)^{(0)},\tilde \pi _q^{(0)}$;
			\State $l = 1$.
			\renewcommand{\algorithmicrequire}{\textbf{Step 1:}}
			\Require
			\State Use $O^l$ and the values of HMM parameters to obtain ${\eta _t}(q)^{(l)}$ and ${\xi _t}(q,p)^{(l)}$ by (13) - (14);
			\State The state transition probability of sequence $O^l$: $a_{q,p}^{(l)} = \frac{{\sum\limits_{t = 1}^T {{\xi _t}(q,p)^{(l)}} }}{{\sum\limits_{t = 1}^{T - 1} {{\eta _t}(q)^{(l)}} }}$;
			\State The state-observation probability of sequence $O^l$: $b_p(j)^{(l)} = \frac{{\sum\limits_{t = 1,{o_t^l} = V(j)}^T {{\eta _t}(p)^{(l)}} }}{{\sum\limits_{t = 1}^T {{\eta _t}(p)^{(l)}} }}$;
			\State The initial state probability of sequence $O^l$: $\pi _q^{(l)} = {\eta _1}(q)^{(l)}$.
			\renewcommand{\algorithmicrequire}{\textbf{Step 2:}}
			\Require
			\State Update $A$ by $\tilde a_{q,p}^{(l)} = \frac{1}{l}\sum\limits_{i = 1}^l {a_{q,p}^{(l)}}$;
			\State Update $B$ by ${{\tilde b}_p}{(j)^{(l)}} = \frac{1}{l}\sum\limits_{i = 1}^l {{b_p}{{(j)}^{(l)}}}$;
			\State Update $\pi$ by $\tilde \pi _q^{(l)} = \frac{1}{l}\sum\limits_{i = 1}^l {\pi _q^{(l)}}$;
			\State $l=l+1$.
			\renewcommand{\algorithmicrequire}{\textbf{Step 3:}}
			\Require
			\State \textbf{IF} $A$, $B$ and $\pi$ converge, or $l = L$ is met, output $\lambda = \{A,B,\pi\}$;\\
			\textbf{ELSE} Return to \textbf{Step 1}.
		\end{algorithmic}
	\end{MYalgorithmic}
\end{figure}
\begin{equation}
{\xi _t}(q,p) = \frac{{{\alpha _t}(q){a_{q,p}}{b_q}({o_{t+1}^l}){\beta _{t + 1}}(p)}}{{\sum\limits_{q = 1}^Q {\sum\limits_{p = 1}^Q {{\alpha _t}(q){a_{q,p}}{b_q}({o_{t+1}^l}){\beta _{t + 1}}(p)} } }}.
\end{equation} \par
As discussed above, when applying $K$-means clustering, the trail of mobility features of vehicle $l$ is turned into one an observation sequence of integers, i.e., ${O^l}$. After that, Baum-Welch algorithm is applied in this paper for the training of HMMs. To ensure the convergence in the training of an HMM ${\lambda _i}$, observations of $L_i$ vehicles are used. For simplicity, the index $i$ is omitted, and the input of the training algorithm is denoted as a set of observations $O = \{ {O^1},...,{O^l},...,{O^L}\}$. The parameters of the HMM are estimated in the iteration of training process. Algorithm 2 gives the detailed procedure of HMM training in the case of discrete characterization of mobility features.

\subsection{HMM training with continuous characterization of mobility features}
The process of clustering in the discrete characterization turns the trails of mobility features into sequences of integers, and then the state-observation probabilities can be written in the form of a matrix, i.e., $B$. However, this may cause a loss of information in continuous data, especially when the number of clusters is relatively small. Gaussian mixture model can be used as an alternative approach to characterize the continuous mobility features in HMM training. The $n$-th column of ${\bf{F}}$ is the vector that includes the trails of features of type $n$ of all $L$ vehicles, which can be denoted as ${\bf{f}}_n$, with the probability density function of $p({{\bf{f}}_n})$. Then, a superposition of $M$ Gaussian distribution is used to fit $p({{\bf{f}}_n })$, i.e.,
\begin{equation}
p({{\bf{f}}_n}) = \sum\limits_{m = 1}^M {\sum\limits_{p = 1}^Q {{\omega _{n,p,m}}p({{\bf{f}}_n}| \mathcal N({\mu _{n,p,m}},{\sigma _{n,p,m}}))} },
\end{equation}
where ${{\omega _{n,p,m}}}$ is the weight that the $p$-th state is modelled by the $m$-th Gaussian component for mobility feature of type $n$, ${{\mu _{n,p,m}}}$ and ${{\sigma _{n,p,m}}}$ are the corresponding mean value and standard deviation of the Gaussian distribution respectively. 
Use a set ${\Theta _n} = \{ {\Omega _n},{{\rm M}_n},{\Sigma _n}\} $ to represent the GMM parameters for the $n$-th type of feature, where ${\Omega _n} = {\rm{\{ }}{\omega _{n,p,m}}{\rm{\} }}$, ${{\rm M}_n} = {\rm{\{ }}{\mu _{n,p,m}}{\rm{\} }}$ and ${\Sigma _n} = {\rm{\{ }}{\sigma _{n,p,m}}{\rm{\} }}$ are three 3-dimensional matrices. Then, we denote $\Theta  = \{ {\Theta _1},{\Theta _2},...,{\Theta _N}\} $ as the set of GMM parameters for the completed $N$ types of mobility features. \par
In the HMM training with continuous characterization of mobility features, maximum likelihood estimation is applied to compute the GMM parameters. The objective is to find the Gaussian distributions that are most likely to fit the probability density functions of the mobility features. For the $n$-th type of feature, it is required to find the $ {\Theta _n}$ which satisfies
\begin{equation}
\arg \mathop {\max }\limits_{{\Theta _n}} p({{\bf{f}}_n}) = \arg \mathop {\max }\limits_{{\Theta _n}} \prod\limits_t^{TL} {p({{\bf{f}}_n}(t))}.
\end{equation}
To obtain the required GMM parameters\cite{8}, an initial guess of ${\Theta}$ is set, and then the probability of the value ${{{\bf{f}}_n}(t)}$ produced by the $m$-th Gaussian component is
\begin{small}
\begin{equation}
{\gamma _{n,p,m,t}} = \frac{{\sum\limits_{p = 1}^Q {{\omega _{n,p,m}} \mathcal N({{\bf{f}}_n}{\rm{(t)|}}{\mu _{n,p,m}},{\sigma _{n,p,m}})} }}{{\sum\limits_{p = 1}^Q {\sum\limits_{j = 1}^M {{\omega _{n,p,j}} \mathcal N({{\bf{f}}_n}{\rm{(t)|}}{\mu _{n,p,j}},{\sigma _{n,p,j}})} } }}.
\end{equation}
\end{small}
The corresponding mean value and standard deviation can be achieved, i.e.,
\begin{equation}
{\mu _{n,p,m}} = \frac{{\sum\limits_{t = 1}^{TL} {{{{{\bf{f}}_n}(t)}}{\gamma _{n,p,m,t}}} }}{{\sum\limits_{t = 1}^{TL} {{\gamma _{n,p,m,t}}} }},
\end{equation}
\begin{equation}
\sigma _{n,p,m}^2 = \frac{{\sum\limits_{t = 1}^{TL} {{{({{{{\bf{f}}_n}(t)}} - {\mu _{n,p,m}})}^2}{\gamma _{n,p,m,t}}} }}{{\sum\limits_{t = 1}^{TL} {{\gamma _{n,p,m,t}}} }}. 
\end{equation}
In this case, the state-observation probabilities are represented by ${\Theta }$ instead of a matrix. Algorithm 3 gives the procedure of HMM training in the continuous characterization of mobility features.
\begin{figure}[!t]
	\begin{MYalgorithmic}
		\algcaption{HMM training with continuous characterization of mobility features}
		\label{alg3}
		\begin{algorithmic}[1]
			\renewcommand{\algorithmicrequire}{\textbf{Input:}}
			\Require
			mobility feature matrix ${\bf{F}}$, number of Gaussian components $M$.
			\renewcommand{\algorithmicrequire}{\textbf{Output:}}
			\Require
			HMM parameters $A$, $\Theta$, and $\pi$.
			\renewcommand{\algorithmicrequire}{\textbf{Initialization:}}
			\Require
			\State Set a initial guess of $A$, $\Theta$, and $\pi$.
			\renewcommand{\algorithmicrequire}{\textbf{Step 1:}}
			\Require
			\State Use the values of ${{\mu _{n,p,m}}}$, ${{\sigma _{n,p,m}}}$, and ${{\omega _{n,p,m}}}$ in ${\Theta}$ to obtain ${\gamma _{n,p,m,t}}$ by (12);
			\State Update the values of ${{\mu _{n,p,m}}}$, ${{\sigma _{n,p,m}}}$, and ${{\omega _{n,p,m}}}$ in ${\Theta}$ according to (13) - (14);
			\State Update $A$ and $\pi$ according to Step 1 and Step 2 in Algorithm 2.
			\renewcommand{\algorithmicrequire}{\textbf{Step 2:}}
			\Require
			\State Return to \textbf{Step 1} and repeat until convergence.
		\end{algorithmic}
	\end{MYalgorithmic}
\end{figure}

\subsection{HMM prediction}
After the HMMs are well trained according to the previous subsection, they can be used to predict the driving intention of a given vehicle. For the targeted vehicle, a historical trail of its mobility features is used to predict its driving intention in following time steps. For example, a historical trail of features may show that the targeted vehicle has been keeping driving in one lane, and this indicates either a trend of lane-change or a lane-keeping behavior in future. It is required to predict which behavior is most likely to happen. \par
Besides, the historical trail of features of the surrounding vehicles in the same time period may also be used together for the driving intention prediction of the targeted vehicle. In this case, the traffic environment around the target vehicle is considered. Since the surrounding vehicles close to the targeted vehicle can have an influence on the driving behaviors of the targeted vehicle, taking the mobility features of them into account can help the models to predict the driving intentions of the targeted vehicle more accurately. However, adding the trails of mobility features of some surrounding vehicles may significantly increase the number of features in the training of HMMs. With a limited maximum number of iterations, the training algorithms may not be well converged in the case of a large mobility feature matrix. As a result, the prediction accuracy may be dropped in some experiments where the algorithms are not well converged. \par
As described above, the historical trail of the mobility features of the targeted vehicle, or the historical trails of the mobility features of the targeted vehicle and the surrounding vehicles, can be used as the observation of the HMM. When predicting the driving intention, the observation is used as the input of the prediction algorithm. Then, the probability that the observation is produced by each HMM ${\lambda _i}$ is calculated by Forward Algorithm \cite{7}. After that, the HMM that are most likely to give this observation is selected, i.e.,
\begin{equation}
i = \mathop {\arg \max }\limits_i P(O|{\lambda _i}),
\end{equation}
where $i \in \{ 1,2,3\}$. In this paper, $i=1$ refers to the driving intention of changing to the left lane, $i=2$ refers to the driving intention of changing to the right lane, and $i=3$ refers to the driving intention of keeping on the original lane. The driving intention that $i$ represents for is considered as the driving intention of the targeted vehicle.

\section{Experimental Results and analysis}
\subsection{Experiment Scenario}
As shown in Fig.3(a), the experiment is implemented with the traffic data of the vehicles on the JianGuoMen Flyover in Beijing. It is a three-layer interconnected overpass, usually with high traffic density and complicated traffic conditions. Given a dataset of historical locations of the vehicles in a period of time on the flyover, the proposed method in this paper is used to predict the driving intentions of the vehicles. To demonstrate the scenario, the dataset is imported into ArcMap and Fig.3(b) is plotted. Each dot is a node on the flyover and the location coordinate of it is available. Each gray line is the central axis of the road segment. The red nodes make up a selected main road on the flyover, which is a two-way road with seven lanes. The historical data of 3910 vehicles on this selected road is collected. The triangles of four different colors represent the trails of four vehicles on this road, and the time when each location is recorded is also shown in the figure. In this paper, the driving intentions on this road are divided into three types, i.e., changing to the left lane, changing to the right lane, and keeping on the original lane. Some of the 3910 pieces of data are used to train the model, while others are used as a test set and gives a prediction accuracy, which is used as criteria for evaluating the performances of the proposed method. \par
In this section, the performances of the proposed driving intention prediction method are evaluated under different conditions. In the experiment, a trail of features of a vehicle is truncated into a sequence of 9 time steps, i.e., $T = 9$. Since each time step is a time period of 0.5 seconds, a trail lasts for 4.5 seconds. A vehicle may keep in one lane or change lanes in the 4.5 seconds. For the convenience of classification, when truncating the mobility data of the vehicles, it is ensured that the lane-changing vehicles complete the lane-change behavior (i.e., move across the lane line) at the last time step of the trail. In addition, a vehicle only changes lane once in one trail.\par
\begin{figure}[!t]
	\centering
	\subfigure[JianGuoMen Flyover in Beijing.]{\includegraphics[width=0.48\textwidth]{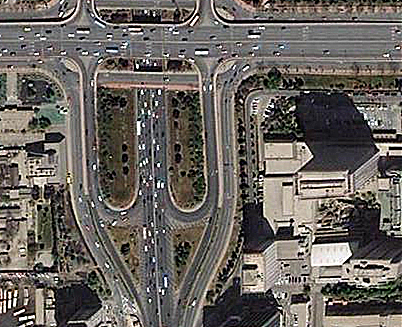}}
	\centering
	\subfigure[Field collected data of vehicles on the selected road.]{\includegraphics[width=0.48\textwidth]{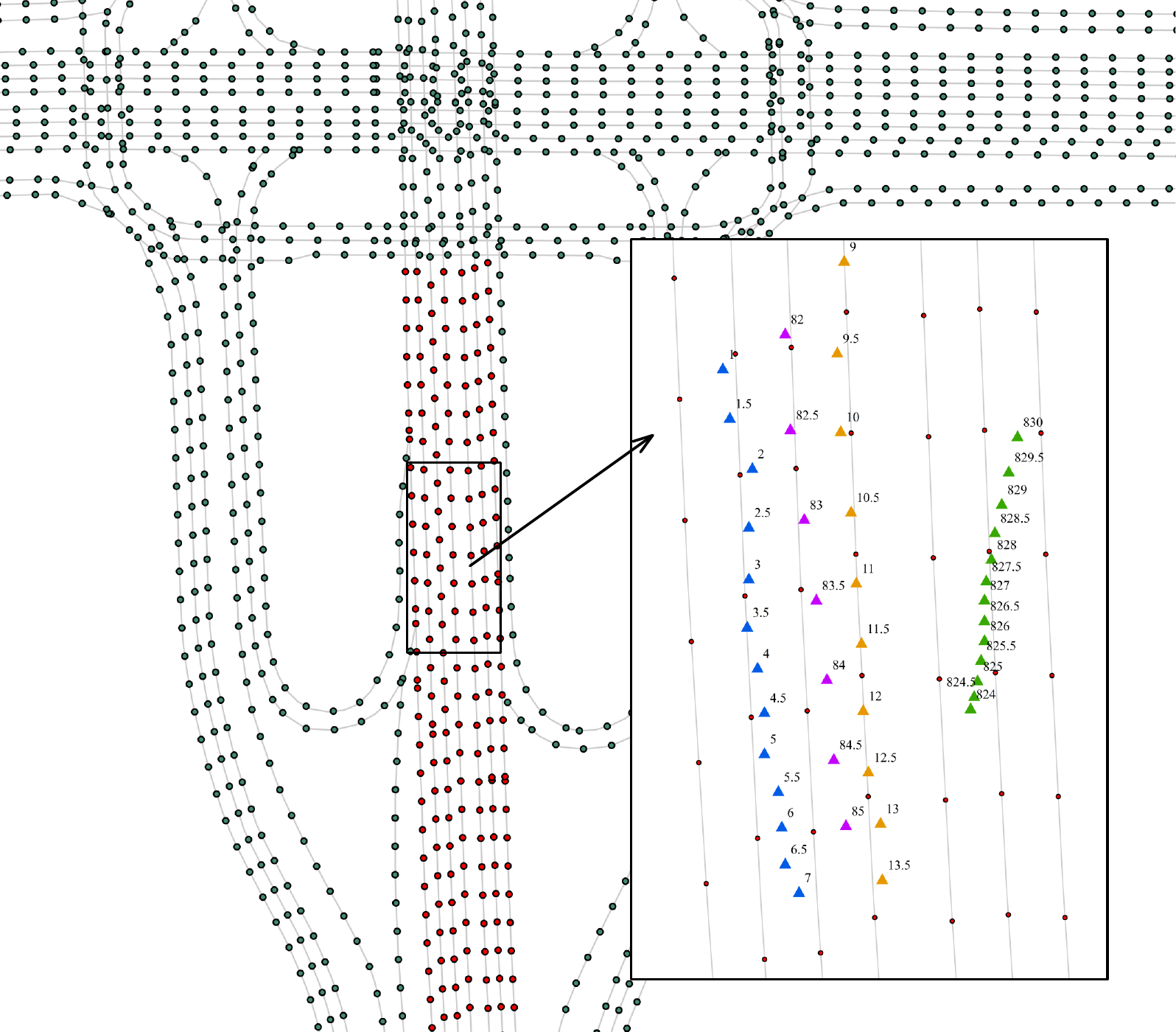}}
	\caption{Simulation scenario: JianGuoMen Flyover and the selected road.}
\end{figure}
As illustrated in Fig.4(a), at each time step, the selected types of mobility features of the targeted vehicle are computed, i.e., the velocity $v$, the moving direction $\theta $, and the offsets from the lane lines $d_1$, $d_2$. The example in Fig.4(b) gives the collected raw data of a vehicle and the values of its selected mobility features. Moreover, since the driving intention of the targeted vehicle may be influenced by the movement of the surrounding vehicles, this paper further takes the trails of mobility features of the surrounding vehicles into account as well. In order to make sure that vehicles in all directions are considered, we choose one nearest surrounding vehicle (if it exists) from each of the 8 regions, instead of simply choosing a specific number of nearest vehicles. The sizes of the regions are determined by $l_1 = 4$ \textit{m}, $l_2 = 6$ \textit{m}, $l_3 = 1$ \textit{m} and $l_4 = 1.5$ \textit{m}. In particular, the values of $l_1$ and $l_2$ depend on how far away the vehicle may have an influence on the targeted vehicle, while $l_3$ and $l_4$ are related to the safe distance between vehicles. Then, the mobility features of the selected surrounding vehicles are added to the elements of the feature matrix ${\bf{F}}$. \par
\begin{figure}[!t]
	\centering
	\subfigure[Mobility features of the targeted vehicle and the selection of the surrounding vehicles.]{\includegraphics[width=0.48\textwidth]{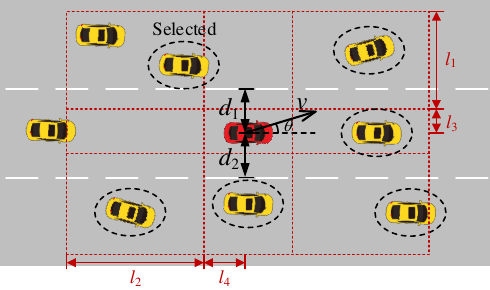}}
	\centering
	\subfigure[An example of the field collected data and the mobility features.]{\includegraphics[width=0.48\textwidth]{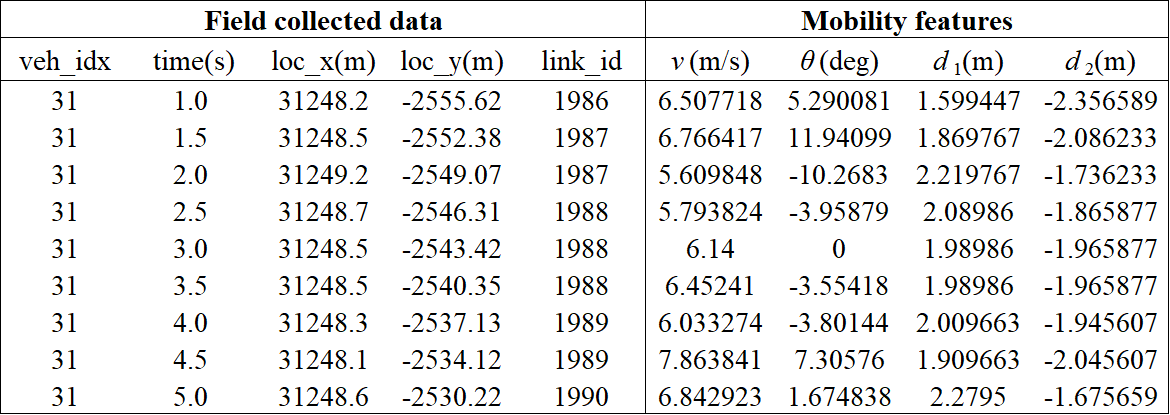}}
	\caption{Selection of mobility features.}
\end{figure}
When evaluating the performance of the prediction, the prediction time refers to the interval between the time that we make the prediction and the time that the trail finishes. For example, when the prediction time equals to 3.0 \textit{s}, it means that to predict the driving intention of the targeted vehicle at $t = T $, we have to make the prediction at 3 seconds earlier. The prediction accuracy is statistically measured. A number of vehicles are used as a test set and the driving intention of them are predicted. The accuracy equals the ratio of correct predictions.  \par

\subsection{Results and Analysis}
Fig.5 shows the performances of the prediction method when different feature characterization approaches with different parameters are used in HMM training and prediction. When the prediction is earlier, the prediction accuracy becomes lower. Moreover, in Fig.5(a), when the discrete characterization of mobility features is applied, a larger number of clusters, i.e., $K$, produces a higher prediction accuracy. Fig.5(b) shows that the continuous characterization of mobility features achieves better performances than the discrete characterization. In Fig.5(c), when the continuous characterization of features is applied, increasing the number of Gaussian components $M$ within a range can improve the performances of driving intention prediction. When the number of hidden states $Q$ becomes larger, the prediction accuracy is improved as well.\par
\begin{figure}[!t]
	\centering
	\subfigure[Prediction accuracy when discrete characterization is applied with different $K$.]{\includegraphics[width=0.48\textwidth]{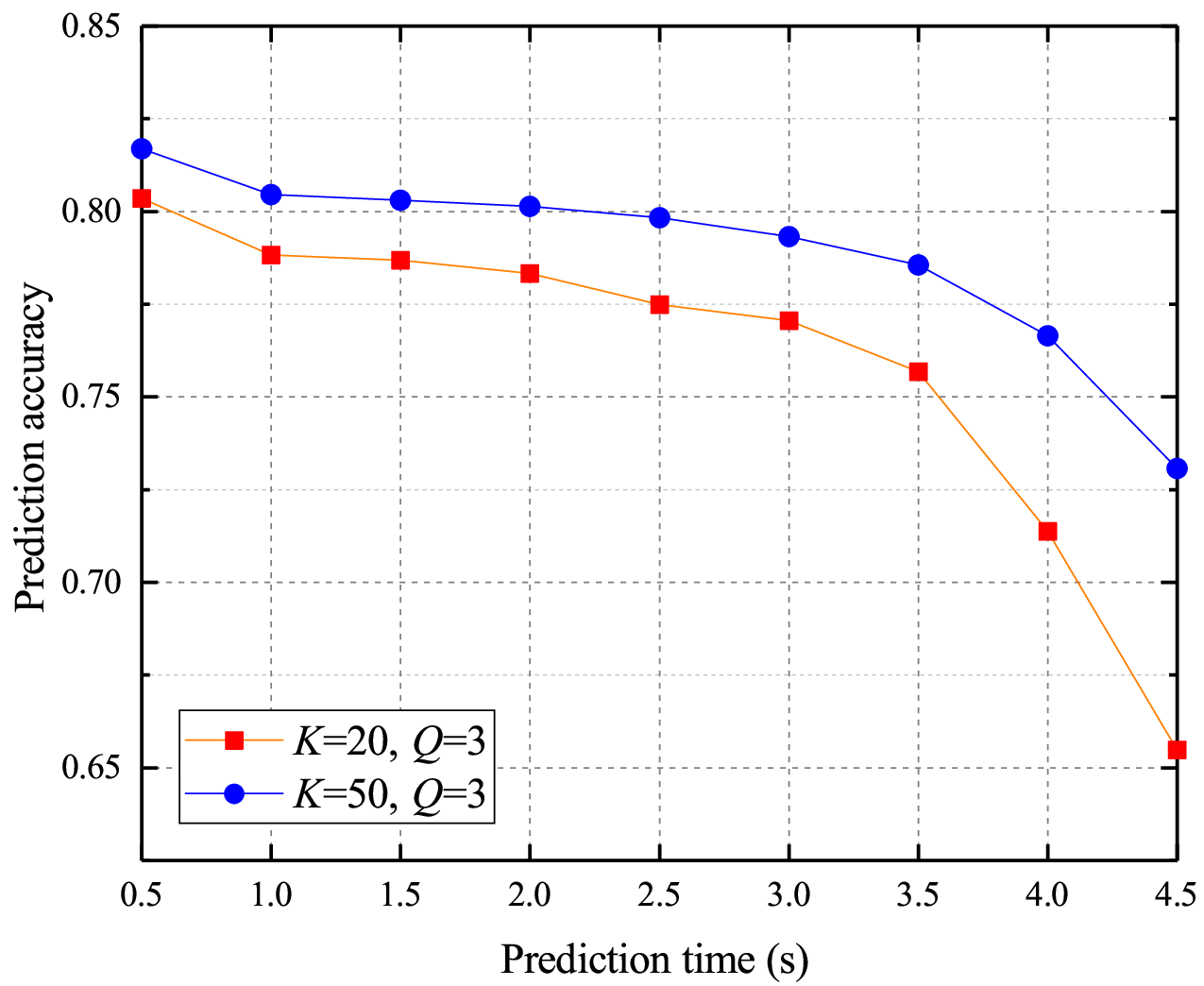}}
	\centering
	\subfigure[Prediction accuracy when discrete characterization or continuous characterization is applied.]{\includegraphics[width=0.48\textwidth]{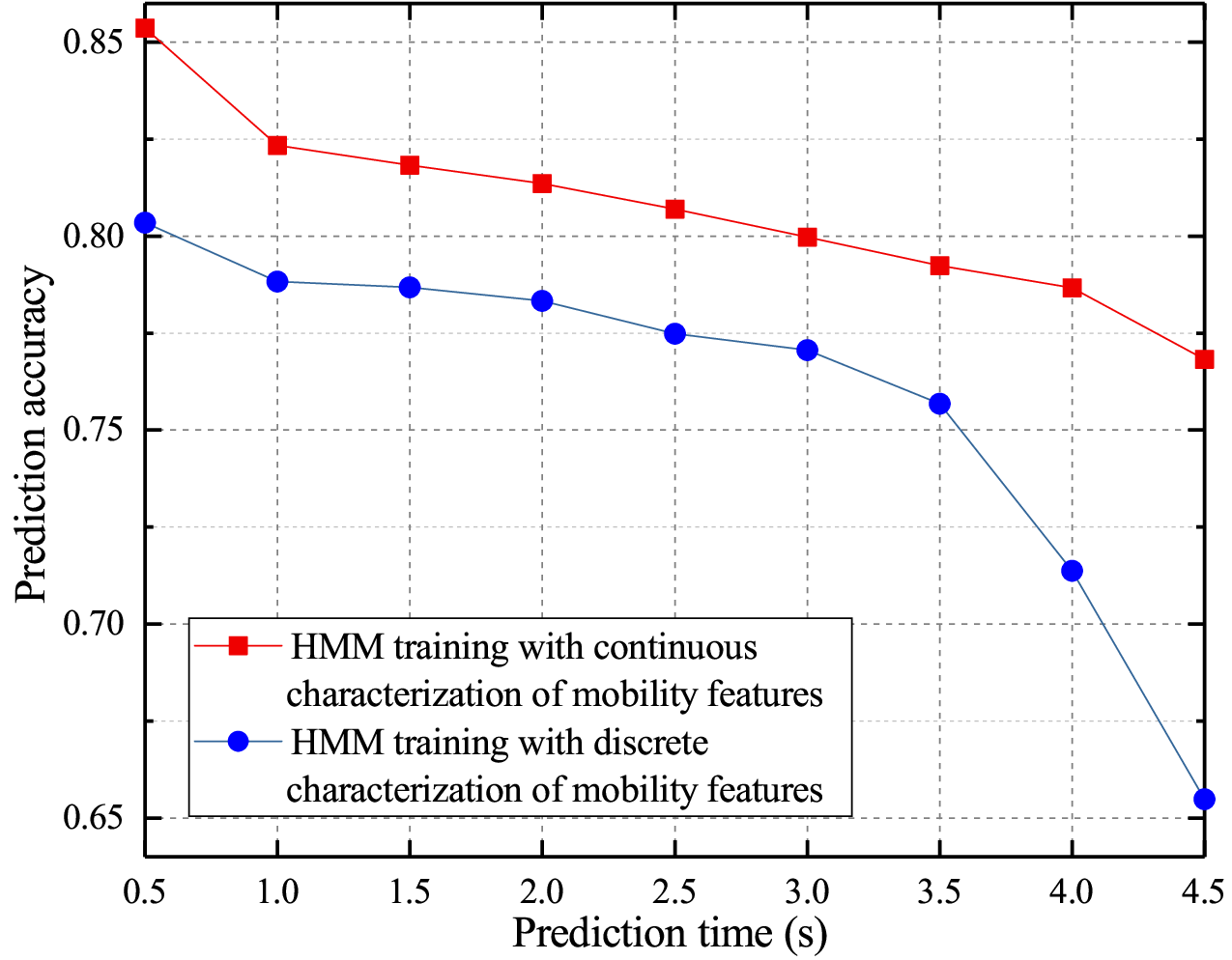}}
	\centering
	\subfigure[Prediction accuracy when continuous characterization is applied with different $M$ and $Q$.]{\includegraphics[width=0.48\textwidth]{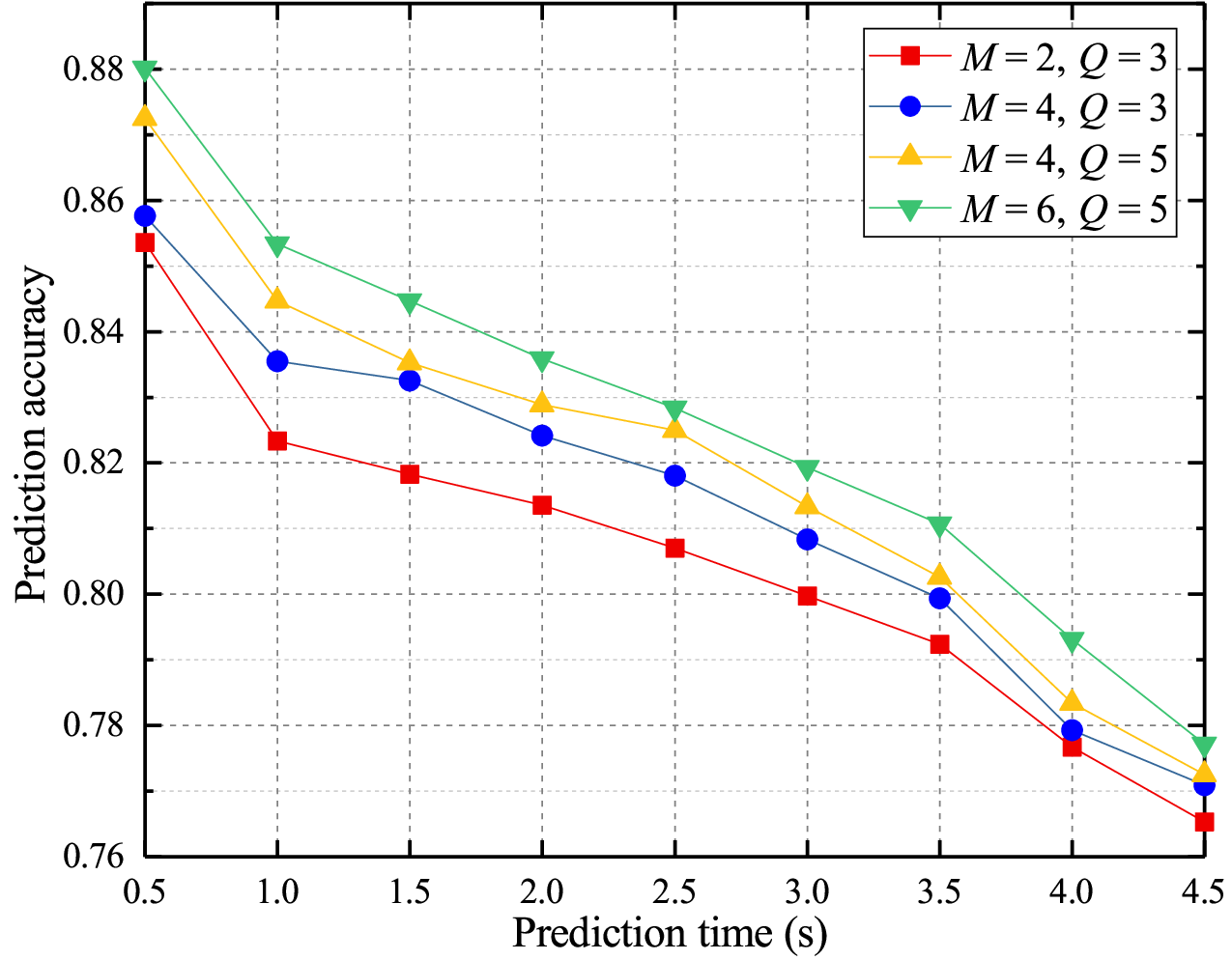}}
	\caption{Prediction accuracy with different characterization
		methods and different parameters.}
\end{figure}
\begin{figure}[!t]
	\centering
	\includegraphics[width=0.48\textwidth]{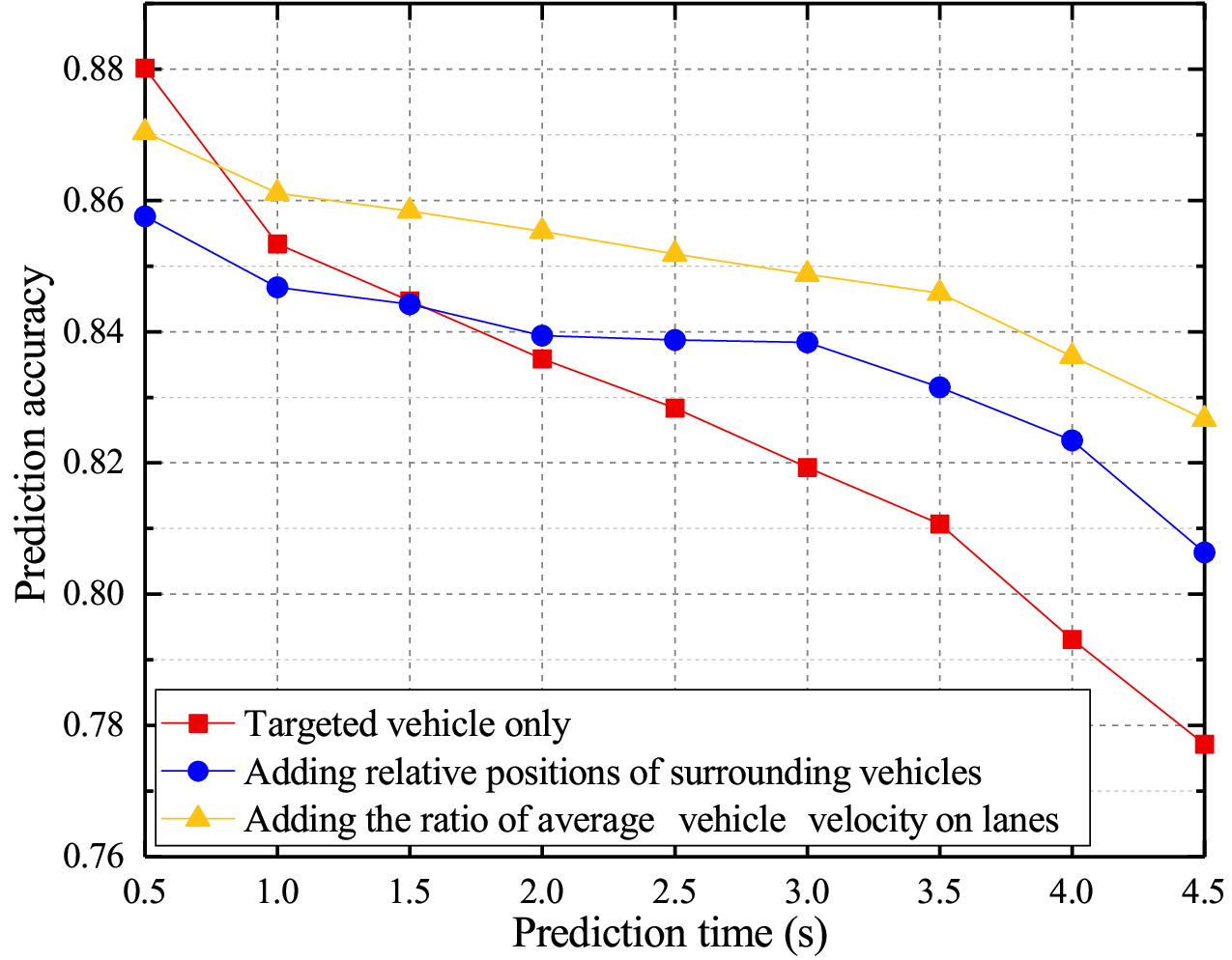}
	\caption{Prediction accuracy different mobility features involved.}
\end{figure}
In Fig.6, the red line with square symbols is obtained when only the mobility features of the targeted vehicle are used as the input of the feature characterization. Then, the mobility features of the surrounding vehicles are taken into account. Since the locations of the surrounding vehicles may limit the driving intention of lane-changing, the relative positions of the selected surrounding vehicles are added to the feature matrix. The relative position of a surrounding vehicle refers to the difference in location coordinates between the surrounding vehicle and the targeted vehicle. As a result, the prediction accuracy is improved when the prediction time is long. After that, since one of the reasons that a driver chooses to change lane is to get an increase in speed, we introduce the ratio of the average velocity of vehicles on the adjacent lane and that of vehicles on the current lane as a mobility features added on the basis of the previous features. The results show that the prediction accuracy is further improved. Note that when the prediction time is smaller than or equals to 1.5 \textit{s}, the prediction accuracy in the case of only mobility features of the targeted vehicle involved is a little higher, compared to the cases where both the mobility features of the targeted vehicle and those of the surrounding vehicles are involved in. This is because in the latter case, the number of the features is significantly increased and the size of the mobility feature matrix is enlarged. As a result, in some experiments, Algorithm 2 or 3 may not be converged after the maximum number of iterations has been reached. The accuracy is a little lower in these experiments, and hence the average prediction accuracy in this case is dropped. In general, the introduction of the mobility features of the surrounding vehicles can increase the accuracy of the prediction of the targeted vehicle.

\section{CONCLUSION}
In this paper, a driving intention prediction method in a mixed-traffic scenario is proposed based on HMM. In the method, either discrete or continuous characterization of the mobility features is applied, and the mobility feature matrix is turned into a set of observations in HMMs. With adequate samples of observations, HMMs representing different driving intentions are trained by the training algorithms. After that, the well-trained HMMs are used to predict the driving intention of a given targeted vehicles. The HMMs are trained and tested with field collected data from a flyover. In the HMM training and prediction, either the mobility features of the targeted vehicle, or the mobility features of the targeted vehicle and the surrounding vehicles are involved in. The numerical results show that the HMMs trained with the continuous characterization of mobility features can give a higher prediction accuracy when they are used for predicting driving intentions. Adopting different parameters in the process of mobility feature characterization gives different performances in prediction. Moreover, when the surrounding vehicles are involved in the training and prediction, the influence of the surrounding traffic on the targeted vehicle is taken into account, and the performances of the prediction method are further improved.


\begin{thebibliography}{99}
\bibitem{5}	
V.~Gadepally, A.~Krishnamurthy, and U.~Ozguner, ``A framework for estimating driver decisions near intersections,"  
\textit{IEEE Transactions on Intelligent Transportion Systems}, vol. 15, no. 2, pp. 637 - 646, Apr. 2014.

\bibitem{16}	
R.~Verma and D.~Vecchio, ``Semiautonomous multivehicle safety,"  
\textit{IEEE Robotics \& Automation Magazine}, vol. 18, no. 3, pp. 44 - 54, Sept. 2011.

\bibitem{10}	
C.~Hubmann, J.~Schulz, M.~Becker, D.~Althoff, and C.~Stiller, ``Automated driving in uncertain environments: planning with interaction and uncertain maneuver prediction,"  
\textit{IEEE Transactions on Intelligent Vehicles}, vol. 3, no. 1, pp. 5 - 17, Jan. 2018.

\bibitem{12} 
K.~Zheng, Q.~Zheng, P.~Chatzimisios, W.~Xiang, and Y~Zhou, `` Heterogeneous vehicular networking: a survey on architecture, challenges, and solutions,"
\textit{IEEE Communications Surveys \& Tutorials}, vol. 17, no. 4, pp. 2377 - 2396, June 2015.

\bibitem{13} 
H.~Liu, H.~Yang, K.~Zheng, and L.~Lei, `` Resource allocation schemes in multi-vehicle cooperation systems,"
\textit{Journal of Communications and Information Networks
}, vol. 2, no. 2, pp. 113 - 125, June 2017.

\bibitem{14} 
Q.~Zheng, K.~Zheng, P.~Chatzimisios, H.~Long, and Fei Liu, `` A novel link allocation method for vehicle-to-vehicle-based relaying networks,"
\textit{Transactions on Emerging Telecommunications Technologies}, vol. 27, no. 1, pp. 64 - 73, Jan. 2016.

\bibitem{9}
J.~Heine, M.~Sylla, T.~Schramm, I.~Langer, B.~Abendroth, and R.~Bruder, ``Algorithm for driver intention detection with fuzzy logic and edit distance," in
\textit{the Proceedings of 2017 IEEE International Conference on Systems, Man, and Cybernetics (SMC)}, pp. 2712 - 2717, Oct. 2017.

\bibitem{2}
Y.~Liang, M.~Reyes, and J.~Lee, ``Real-time detection of driver cognitive distraction using support vector machines,"
\textit{IEEE Transactions on Intelligent Transportation Systems}, vol. 8, no. 2, pp. 340 - 350, June 2007.

\bibitem{11}
Y.~Liao, S.~Li, G.~Li, W.~Wang, B.~Cheng, and F.~Chen, ``Detection of driver cognitive distraction: An SVM based real-time algorithm and its comparison study in typical driving scenarios," in 
\textit{the Proceedings of 2016 IEEE Intelligent Vehicles Symposium (IV)}, pp. 394 - 399, June 2016.

\bibitem{3}
Y.~Kishimoto and K.~Oguri, ``A Modeling Method for Predicting Driving Behavior Concerning with Driver’s Past Movements," in 
\textit{the Proceedings of 2008 IEEE International Conference on Vehicular Electronics and Safety}, pp. 132 - 136, Sept. 2008.

\bibitem{1}
S.~Amsalu, A.~Homaifar, A.~Karimoddini, and A.~Kurt, ``Driver intention estimation via discrete hidden Markov model," in
\textit{the Proceedings of 2017 IEEE International Conference on Systems, Man, and Cybernetics (SMC)}, pp. 2712 - 2717, Oct. 2017.

\bibitem{4}
K.~Yamada, H.~Matsuyama, and K.~Uchida, ``A method for analyzing interaction of driver intention through vehicle behavior when merging," in 
\textit{the Proceedings of 2014 IEEE Intelligent Vehicles Symposium}, pp. 158 - 163, June 2014.

\bibitem{6}	
T.~Kanungo, D.~Mount, N.~Netanyahu, C.~Piatko, R.~Silverman, and A.~Wu, ``An efficient k-means clustering algorithm: analysis and implementation,"
\textit{IEEE Transactions on Pattern Analysis and Machine Intelligence}, vol. 27, no. 4, pp. 881 - 892, Aug. 2002.

\bibitem{17}
K.~Krishna and M.~Murty, ``Genetic K-means algorithm,"
\textit{IEEE Transactions on Systems, Man, and Cybernetics, Part B (Cybernetics)}, vol. 29, no. 3, pp. 433 - 439, June 1999.

\bibitem{7} 
L.~Rabiner and B.~Juang, ``An introduction to hidden Markov models,"
\textit{IEEE ASSP Magazine}, vol. 3, no. 1, pp. 4 - 16, Jan. 1986.

\bibitem{15}
L.~Rabiner, `A tutorial on hidden Markov models and selected applications
in speech recognition,"
\textit{Proceedings of the IEEE}, vol. 77, no. 2, pp. 257 - 286, Feb. 1989.

\bibitem{8} 
G.~Xuan, W.~Zhang, and P.~Chai, ``EM algorithms of Gaussian mixture model and hidden Markov model," in
\textit{the Proceedings of 2001 International Conference on Image Processing}, pp. 145 - 148, Oct. 2001.


\end{thebibliography}
\end{document}